%% file: main.tex
\documentclass[journal]{IEEEtran}

\usepackage{amsmath,amsfonts}
\usepackage{algorithmic}
\usepackage{array}
\usepackage[caption=false,font=normalsize,
   labelfont=sf,textfont=sf]{subfig}
\usepackage{textcomp}
\usepackage{stfloats}
\usepackage{url}
\usepackage{verbatim}
\usepackage{graphicx}
\usepackage{balance}

\usepackage[style=ieee,natbib=true]{biblatex}
\usepackage[ruled,vlined,linesnumbered]{algorithm2e}
\SetKw{Break}{break}

\usepackage{williamstyle}
\usepackage{xr}

\newrobustcmd\best{\DeclareFontSeriesDefault[rm]{bf}{b}\bfseries}

\begin{document}

\title{Graph-Dictionary Signal Model for Sparse Representation of Multivariate Data}

\author{ William Cappelletti, Pascal Frossard \\ LTS4, EPFL,
Lausanne, Switzerland}

\maketitle

\begin{abstract}
  \input{sections/abstract.tex}
\end{abstract}


\input{sections/introduction.tex}
\input{sections/related.tex}
\input{sections/graph-signal-model.tex}
\input{sections/algorithm.tex}
\input{sections/experiment_model.tex}
\input{sections/experiments.tex}
\input{sections/conclusion.tex}

\section*{Acknowledgments}
\input{sections/acknowledgments.tex}

\printbibliography



\end{document}

%% file: sections/abstract.tex
Representing and exploiting multivariate signals requires capturing relations between variables, which we can represent by graphs. Graph dictionaries allow to describe complex relational information as a sparse sum of simpler structures, but no prior model exists to infer such underlying structure elements from data. We define a novel Graph-Dictionary signal model, where a finite set of graphs characterizes relationships in data distribution as filters on the weighted sum of their Laplacians. We propose a framework to infer the graph dictionary representation from observed node signals, which allows to include a priori knowledge about signal properties, and about underlying graphs and their coefficients. We introduce a bilinear generalization of the primal-dual splitting algorithm to solve the learning problem. We show the capability of our method to reconstruct graphs from signals in multiple synthetic settings, where our model outperforms popular baselines. Then, we exploit graph-dictionary representations in an illustrative motor imagery decoding task on brain activity data, where we classify imagined motion better than standard methods relying on many more features. Our graph-dictionary model bridges a gap between sparse representations of multivariate data and a structured decomposition of sample-varying relationships into a sparse combination of elementary graph atoms.

%% file: sections/introduction.tex
\section{Introduction}\label{sec:intro}

Multivariate signals represent joint measurements from multiple sources, and commonly arise in different fields.
They might represent electrical potentials generated by brain activity, as well as stock values in the financial market, temperatures across weather stations, or traffic measurements at road junctions.
Generally, we study these signals to characterize the system on which we measure them; for instance, to classify brain activity for motor imagery analysis, or to capture hidden relationships between variables.
We are interested in effective representations that can provide an explanation of the relational information between different sample variables, to identify the role of each source, and describe the distribution of their outcomes.
Graphs provide a natural representation of structured data with pairwise relationships, described by \emph{edges} between \emph{nodes}, however, these relationships are often hidden, and we must infer them from data.

\begin{figure}[t]
  \centering
  \includegraphics[width=\linewidth]{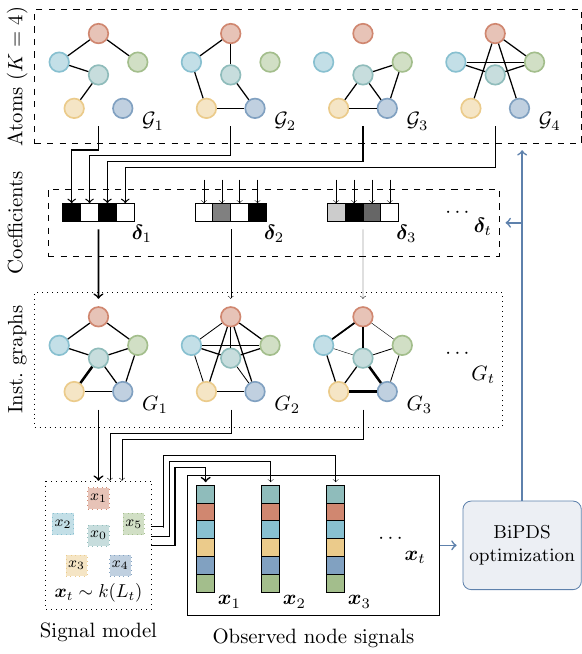}
  \caption{\label{fig:graph-dict}
    Representation of our novel graph-dictionary signal model.
    Solid black arrows illustrate the generative process, where coefficients $\bm \delta_t$ mix atoms $\mathcal G_1, \dots, \mathcal G_K$ from the dictionary to give instantaneous graphs $G_t$, whose Laplacians $L_t$ characterize the signal model from which signals $\bm x_t$ are sampled. In particular, the \textbf{thick} arrows follow the generation of $\bm x_1$.
    Along {\color{nord10} blue lines} we see the inverse problem, which jointly learns atoms and coefficients from observed signals using our Bilinear Primal Dual Splitting algorithm.
  }
\end{figure}

Many multivariate systems are governed by complex instantaneous dynamics, which can vary for every data sample, but which might be modeled as the joint effect of simpler structures.
For instance, when measuring the electrical brain activity of a subject, we capture ionic currents resulting from multiple synaptic processes happening at the same time---like vision processing and motor activity.
Furthermore, the contribution of each simpler dynamic is often sparse in time; for example on traffic measurements, where commuters drive during specific hours on their usual path, which might overlap with the mobility patterns of service vehicles, or tourist cars.

We approach this representation learning problem through \emph{dictionary learning} \citep{rubinsteinDictionariesSparseRepresentation2010}.
This framework aims to recover a finite set of atoms, called a \emph{dictionary}, that can encode data through sparse coefficients.
We introduce a novel Graph-Dictionary signal model, \emph{GraphDict} in short, to jointly address the representation and the structure learning problems.
Following the \emph{Graph Structure Learning} literature \citep{dongLearningGraphsData2019,mateosConnectingDotsIdentifying2019}, we use signal models to link each $N$-variate signal sample~$\bm x_t \in \R^N$ to an underlying \emph{instantaneous} graph~$G_t$.
Our novel approach, illustrated in \cref{fig:graph-dict}, uses a dictionary of graphs to characterize each $G_t$ as the linear combination of $K$ graph \emph{atoms} $\G_1, \dots, \G_K$ with respective \emph{coefficients} $\bm \delta_t \in [0,1]^K$.
The coefficient vector corresponding to each sample provides an embedding which is inherently explainable.
More precisely, each entry describes the contribution of the corresponding graph atom, with edges defining instantaneous relationships that generate the observed signal.
Our objective function jointly characterizes atoms and coefficients, and its optimization requires a novel generalization of the Primal-Dual Splitting algorithm, which we define by leveraging the adjoint functions of a bilinear operator \citep{arensAdjointBilinearOperation1951}.

We present a series of experiments to test the capabilities of our model.
In two experiments on synthetic data, we show that GraphDict outperforms literature baselines in reconstructing instantaneous graphs from signals.
Finally, we exploit the sparse representations given by GraphDict coefficients in an exploratory downstream classification task consisting in motor imagery decoding on brain activity data. We show that with few features on three connectivity atoms identified by our model, we classify imagined motion better than brain-state models from the neuroscience literature.


Our main contributions are the following:
\begin{itemize}
  \item \textbf{Graph signal model.} We introduce \emph{GraphDict}, a novel probabilistic model that defines multivariate signal distributions with graph dictionaries. We formulate signal representation learning as a maximum a posteriori estimate which allows for domain-specific priors on both the graph structures and the contributions of each dictionary atom.
  \item \textbf{Optimization algorithm.} We propose a novel bilinear generalization of the Primal-Dual Splitting algorithm to jointly learn edge weights of dictionary atoms and their mixing coefficients.
  \item \textbf{Practical implementation.} We illustrate some specific loss functions that include common hypotheses from graph structure learning and sparse representations, and we derive their optimization steps.
  \item \textbf{Performance Evaluation.} We present three experiments to investigate the ability of GraphDict's declinations to reconstruct instantaneous graphs and provide helpful and explainable representations.
  Our model can reliably reconstruct instantaneous graphs on synthetic data, and its sparse coefficients' representation outperforms established methods in an exploratory downstream classification task on motor imagery decoding.
\end{itemize}

The proposed graph-dictionary signal model provides a flexible tool to reconstruct relationships between variables in multivariate data, and decompose them in atomic graphs. The sparse representation corresponding to the network-atoms coefficients is useful in downstream tasks, and explicitly relates feature learning to relational information.

%% file: sections/related.tex
\section{Related Work}\label{sec:related}

Learning graph dictionaries is a very recent trend, and multiple works \citep{vincent-cuazOnlineGraphDictionary2021,liuRobustGraphDictionary2022, zengGenerativeGraphDictionary2023, dasTensorGraphDecomposition2024} address this problem starting from actual graph observations, while our framework only requires node signals.
Methods based on optimal transport use different graph distances to learn dictionaries from unaligned graphs. \citet{vincent-cuazOnlineGraphDictionary2021} propose an online learning method based on Gromov-Wasserstein (GW) Discrepancy, and \citet{liuRobustGraphDictionary2022} expand this framework to be more robust to graph perturbations.
\citet{zengGenerativeGraphDictionary2023} propose a non-linear graph dictionary learning algorithm, and they formulate the problem from a generative perspective with a Fused GW Mixture Model, which combines optimal transport and kernel estimation.
\citet{dasTensorGraphDecomposition2024} use tensor decomposition to identify a limited number of \emph{key mode graphs}, equivalent to dictionary atoms, that can explain a temporal network, and whose linear combination represents its evolution.
Furthermore, they use an alternating algorithm to optimize the mode graphs and the temporal factor variables, which may be suboptimal

To the best of our knowledge, the only previous work that links signals to graph dictionaries is from \citet{thanouParametricDictionaryLearning2013}, who suggest learning dictionaries of structured graph kernels from signals.
They observe signals over a predefined graph, and the atoms of their dictionary represent different graph filters on the edges of such network.
Our work, on the other hand, does not define the atoms as subgraphs of a known network, but learns edges from scratch, resulting in greater flexibility to capture complex dynamics.
Still, our probabilistic approach allows introducing further regularization terms, and in \cref{sec:experiments} we explain how to encompass the settings of \citet{thanouParametricDictionaryLearning2013}.
The recent paper from \citet{karaaslanliMultiviewGraphLearning2025} addresses a similar problem as they infer a multiview graph model from node signals, where each view is a perturbed version of an underlying consensus graph.
Compared to our graph dictionary formulation, they learn a single recurrent structure---namely the consensus graph---of which views are \enquote{small} perturbations, while we allow more flexibility in the topologies on which signals are observed by inferring multiple atoms, which provide instantaneous graphs through linear combinations.

Another line of work that we cover with our Graph-Dictionary signal model consists in signal clustering and graph learning.
In fact, we can see each cluster as an atom of a dictionary, with signal coefficients providing a vector encoding of cluster assignments.
\citet{mareticGraphHeatMixture2018}, and \citet{mareticGraphLaplacianMixture2020} propose Graph Laplacian Mixture Models (GLMM) in which precision matrices of the signals' distribution depend on some function of the graph Laplacians. Similarly, \citet{araghiGraphsAlgorithmGraph2019} design an algorithm for signals clustering and graph learning called $K$-Graphs, which refines assignments and graph estimation by expectation-maximisation.
Both methods rely on a Gaussian assumption on the signal distribution, but while the latter model enforces hard assignments, like the $K$-means algorithm, GLMM performs soft clustering, by estimating the probability that a sample belongs to each cluster.
Still, clustering methods need a significant separation between the underlying distributions; in graph signal models, this means they might struggle to discriminate samples if the underlying networks share multiple edges.
Furthermore, they cannot capture continuous changes which can arise in more structured data such as multivariate time series.

%% file: sections/graph-signal-model.tex
\section{Graph-Dictionary Signal Model}\label{sec:graph-signal-model}

In this section, we introduce \emph{GraphDict}, our novel Graph-Dictionary signal model, and we describe the corresponding representation learning problem.


\subsection{Graphs and Signals}\label{ssec:graph-signal-pre}

Our data $\bm X \in \R^{T \times N}$ represent a set of $T$ signals of dimension $N$. For each $t \in T$ we observe a multivariate signal $\bm x_t \in \R^N$  on the nodes of a graph $G_t = (N, \bm w_t)$ whose edges $E \subset N \times N$ and corresponding weights $\bm w_t \in \R_+^{E}$ are unknown.
Each edge $(i,j)_t$ describes a relation between signal outcomes $x_{ti}$ and $x_{tj}$ on nodes $i$ and $j$ for sample $t$, with a strength proportional to its weight~$w_{tij}$.
We focus on undirected graphs with nonnegative weights,
which we represent as vectors $\bm w \in \R_+^{N (N-1) / 2} = \R_+^{E}$.

Graph Signal Processing \citep{ortegaIntroductionGraphSignal2022} provides us tools to understand the link between signals and the networks on which we observe them.
One of its most important tools is the \emph{combinatorial Laplacian}~$\bm L$ of a graph, which is defined as the difference between the diagonal matrix of node degrees and the adjacency matrix.
Expressing it as a function of vectorized weights $\bm w$, the Laplacian  is a linear operator from $\bm \R_+^{E} \to \R^{N \times N}$:
\begin{equation}
  [ \bm L_{\bm w} ]_{ij} = \begin{cases}
      - w_{(i,j)} & i \neq j, (i,j) \in E, \\
      \sum_{k: (i, k) \in E} w_{(i,k)} & i = j, \\
      0 & \text{otherwise} .
    \end{cases}
\end{equation}

Of particular interest is the eigendecomposition of the Laplacian $\bm L = \bm U \bm \Lambda \bm U^\top$, with $\bm \Lambda$ the diagonal matrix of eigenvalues $\lambda_1 = 0 \leq \ldots \leq \lambda_N$.
We define the \emph{Graph Fourier Transform} of a signal $\bm x$ as the projection on the eigenbasis $\hat{\bm x} = \bm{U x}$, and noting that eigenvectors are invariant signals, we introduce \emph{graph frequencies} as the corresponding eigenvalues.
We can finally define graph filters $k$ as functions acting in the graph frequency domain \citep{sandryhailaDiscreteSignalProcessing2013}, and represent them as
\begin{equation}
  k(\bm L)=\bm U k(\bm \Lambda)\bm U^\top
  .
\end{equation}


\subsection{Probabilistic Signal Model}\label{ssec:graph-signal-prob}

We propose a model for multivariate data where the state of the system at each sample~$t$ is described by an underlying \emph{instantaneous} graph~$G_t$, given by a linear combination of $K$ networks, which constitute the \emph{atoms} of a \emph{dictionary} of graphs.
Our underlying hypothesis is that many complex systems can be described as the superposition of simpler recurring patterns.

Each $N$-variate sample signal $\bm x_t \in \R^N$ arises from a graph filter~$k(\bm L_t)$ defined on the instantaneous Laplacian $\bm L_t$, applied to a random vector $\bm \eta_t \in \R^{N}$.
This formulation provides much freedom to model complex systems, since by fixing $k$ we are able to focus on specific properties that the signals $\bm x_t$ display on the graph $G_t$, while allowing for great flexibility in the topology of the latter.
For instance, we can consider iid multivariate white noise~${\bm \eta_t \sim \mathcal N (\bm 0, \Id_N)}$, so that
\begin{equation}\label{eq:noise-filter}
  \bm x_t = k(\bm L_t) \bm \eta_t \sim \mathcal N \left( \bm 0, k^2(\bm L_t) \right)
  .
\end{equation}

We define each instantaneous graph $G_t$ through coefficients $\bm \delta_t \in [0,1]^K$ and edge weights of each atom $\bm w_k \in \R_+^{E}$.
We jointly represent the graph dictionary as the matrix ${\bm W = [\bm w_1^\top, \dots, \bm w_K^\top]} \in \R_+^{K \times E}$, whose rows identify atoms, and we express instantaneous Laplacians as a bilinear form of coefficients $\bm \delta_t$ and weights $\bm W$ as follows
\begin{equation}\label{eq:additive-lapl}
  \bm L_t = \bm L(\bm \delta_t, \bm W)
    = \sum_k \delta_{tk} \bm L_{\bm w_k}
    \in \R^{N \times N}
    .
\end{equation}

Conditionally to this graph-dictionary formulation, the signal distribution $P_x$ is characterized by $\bm \delta_t$ and $\bm W$.
In this work, we assume that atom weights and coefficients are independent, meaning that structure and appearance do not influence each other.
In this way, we can split the joint PDF in
\begin{equation}\label{eq:pdf-full}
  P \left(\bm X, \bm W, \bm \Delta \right)
    = P_w(\bm W) P_c(\bm \Delta) P_x \left( \bm X \mid \bm W, \bm \Delta \right)
    ,
\end{equation}
where $P_w$ and $P_c$ are the prior distributions of weights and coefficients respectively, and we gather all coefficients $\bm \delta_t$ in the rows of the matrix ${\bm \Delta
\in [0,1]^{T \times K}}$.


\subsection{Representation Learning Problem}\label{ssec:representation-learning}

With the characterization of signal distribution with the graph dictionary and coefficients from \cref{eq:pdf-full}, we can define the representation learning problem as a maximum a posteriori (MAP) objective.
Observing data $\bm X \in \R^{T \times N}$, we aim to find graph-dictionary weights~$\bm W$ and coefficients~$\bm \Delta$ as the minimizers of the corresponding negative log-likelihood
\begin{equation}\label{eq:objective-nll}
  \underset{\bm W, \bm \Delta}{\arg\min}\
    \ell_w(\bm W) + \ell_c (\bm \Delta)
    + \ell_x(\bm X, \bm W, \bm \Delta))
    ,
\end{equation}
where $\ell_w$, $\ell_c$, and $\ell_x$ are the negative logarithms of $P_w$, $P_c$, and~$P_x$ respectively.

This formulation leaves great flexibility for $\ell_w$, $\ell_c$, and $\ell_x$ which can enforce structure and domain-specific properties on atoms weights, coefficients, and signals respectively.
While such priors are driven by the application, in \cref{sec:algo} we present a general algorithm to solve \cref{eq:objective-nll} for $\bm W$ and $\bm \Delta$.
In \cref{sec:models}, we illustrate a practical example based on common assumptions, that we use in our experiments of \cref{sec:experiments}.

%% file: sections/algorithm.tex
\section{Bilinear Primal-Dual Splitting Algorithm}\label{sec:algo}

\input{sections/content/algo_bi_pds.tex}

To find a solution to the MAP objective~\eqref{eq:objective-nll} we propose \cref{algo:bi-pds}, a generalization of the Primal-Dual Splitting (PDS) algorithm by \citet{condatPrimalDualSplitting2013}.
Supposing that we can rearrange $\ell_w, \ell_c$ from \cref{eq:objective-nll} into a sum of Lipschitz-differentiable and proximable functions---denoted by $f$ and $g$ respectively, and separable in $\bm W$ and $\bm \Delta$---we can rewrite our optimization problem as
\begin{equation}\label{eq:objective-nll-split}
  \underset{\bm W, \bm \Delta}{\arg\min}\
    f(\bm W, \bm \Delta)
    + g(\bm W, \bm \Delta)
    + h \left( \bmr L(\bm \Delta, \bm W) \right)
    ,
\end{equation}
where $h$ incorporates the signal log-likelihood and the graph filter from \cref{eq:noise-filter}, which we apply to the tensor $\bmr L(\bm \Delta, \bm W) \in \R^{T \times N \times N}$ that stacks instantaneous Laplacians.

The original PDS is a first-order algorithm to solve minimization problems of the following form
\begin{equation}\label{eq:objective-pds}
  \underset{\bm x \in \R^N}{\arg\min} \quad f(\bm x) + g(\bm x) + h(\bm{A x}),
\end{equation}
which, except for the last term, corresponds to our objective in \cref{eq:objective-nll-split}.
The limitation of PDS is that ${\bm A: \R^N \to \R^M}$ has to be a linear operator on~$\bm x$, while our $\bmr L$ is a bilinear operator combining $\bm \Delta$ and $\bm W$, for which we need to design a specific method.

PDS looks for solutions to both a primal and a dual problem, given respectively by $\bm x$ and $\bm{Ax}$, and iteratively brings them closer by combining gradient and proximal descent \citep{parikhProximalAlgorithms2013}.
More precisely, given proximal parameters $\tau, \sigma > 0$, and an initial estimate $(\bm x_0, \bm y_0) \in \R^N \times \R^M$, it performs the following iteration for every $n \geq 0$, until convergence:
\begin{subequations}\label{eq:pds-original}
\begin{align}
  \bm x_{n+1} & \gets
    \prox_{\tau g}\left( \bm x_n - \tau \left( \nabla f(\bm x_n) + \bm A^* \bm y_n \right) \right)
    , \label{eq:pds-primal} \\
  \bm y_{n+1} &
    \gets \prox_{\sigma h^*} \left( \bm y_n + \sigma \bm A \left( 2 \bm x_{n+1} - \bm x_n \right) \right)
    , \label{eq:pds-dual}
\end{align}
\end{subequations}
where $\bm A^*$ is the adjoint of $\bm A$, and $h^*$ is the Fenchel conjugate of the function $h$.

The main challenge of our setting comes from optimizing jointly for $\bm W$ and $\bm \Delta$ the function $h$, composed with the graph filter $k$.
With the separation and parallel update of the primal variables $\bm W_n$ and $\bm \Delta_n$, we define Bilinear Primal-Dual Splitting in \cref{algo:bi-pds}.
We note that the update of the dual variable $\bm Y_n$ in \cref{eq:bi-pds-dual} corresponds to the one in \cref{eq:pds-dual}, with operator $\bm A$ substituted by the bilinear one~$\bmr L$.

To generalize PDS, we define the dual variable using the bilinear form ${\bmr Y = \bmr L(\bm \Delta, \bm W)}$.
Then, we observe that the primal update in \cref{eq:pds-original} relies on the adjoint of the linear operator $\bm A^*$, which maps the dual variable to the space of the primal one.
In \cref{eq:objective-nll-split}, supposing that $g$ and $f$ are separable in the primal variables $\bm W$ and $\bm \Delta$, we can define two update steps, each using the partial adjoints of~$\bmr L$ \citep{arensAdjointBilinearOperation1951}, that we show to be
\begin{equation}
  \begin{aligned}
    \bmr L^*(\bm Y, \bm W) &= \bm W d \bmr Y^\top ,\\
    \bmr L^{**}(\bm \Delta, \bm Y) &= \bm \Delta^\top d \bmr Y ,
  \end{aligned}
  \label{eq:adjoint-operators}
\end{equation}
where $d \bmr Y \in \R^{T \times E}$ is the tensor whose entries are given by
\begin{equation}
\big[ d \bmr Y \big]_{t e_{nm}}
  = Y_{tnn} + Y_{tmm} - Y_{tnm} - Y_{tmn} .
\end{equation}

In the notation from \citet{arensAdjointBilinearOperation1951}, we let $A, B, C$ be normed linear spaces on $\R$, and we denote their dual and double dual spaces by $.^*$ and $.^{**}$ respectively.
Given a bilinear operator $\bm L : A \times B \to C$, its adjoints should satisfy
\begin{align}
  \bm L^* : C^* \times A & \to B^* , \nonumber \\
    \bm L^*(\bm f, \bm x)(\bm y) &= \bm f(\bm L(\bm x, \bm y)) ;
    \label{eq:adjoint-1} \\
  \bm L^{**} : B^{**} \times C^* & \to A^* , \nonumber \\
      \bm L^{**}(\bm y, \bm f)(\bm x) &= \bm f(\bm L(\bm x, \bm y)) .
  \label{eq:adjoint-2}
\end{align}

In our case, the bilinear operator $\bmr L : \mathcal{A} \times \mathcal{W} \to \mathcal{L}_T$
is defined as
\begin{equation}
  \begin{aligned}
  \left[ \bmr L(\bm \Delta, \bm W) \right]_{tnm} &= \left[ \bmr L_{\bm \Delta^\top \bm W} \right]_{tnm} \\
  &= \begin{cases}
      - \sum_k \delta_{kt} w_{k (n,m)} &  n \neq m, \\
      \sum_k \delta_{kt} \sum_l w_{k (n,l)} & n = m,
    \end{cases}
  \end{aligned}
\end{equation}
and it maps from the spaces of atom coefficients $\mathcal{A} = [0,1]^{K \times T}$ and graph weights $\mathcal W = \R_+^{K \times E}$ to that of instantaneous Laplacians $\mathcal{L}_T \subset \R^{T \times N \times N}$.

We obtain the adjoint operators by solving the identities~\ref{eq:adjoint-1},~\ref{eq:adjoint-2} for the standard dot products of the given spaces.
Starting from the adjoint operator with respect to weights, $\bmr L^{*}: \R^{T \times N \times N} \times [0,1]^{K \times T} \to \R_+^{K \times E}$, we observe

\begin{subequations}\label{eq:op_adj_weights}
  \allowdisplaybreaks
  \begin{align}
    \big\langle \bmr Y, &\,
      \bmr L( \bm \Delta, \bm W) \big\rangle
      = \sum_{tnm} Y_{tnm} \left[ \bmr L_{\bm \Delta^\top \bm W} \right]_{tnm} \\
    &= \begin{multlined}[t]
      \sum_{tnm, n \neq m} Y_{tnm} \sum_{k} \delta_{kt}( -w_{k (n,m)}) \\
        + \sum_{tn} Y_{tnn} \sum_{k} \delta_{kt} \sum_{l} w_{k (n,l)}
    \end{multlined} \\
    &= \begin{multlined}[t]
      \sum_{\substack{k(n,m) \\ n < m }} w_{k (n,m)} \left(
          - \sum_t \delta_{kt} \left( Y_{tnm} + Y_{tmn} \right)
        \right) \\
        + \sum_{k (n,l)} w_{k(n,l)} \sum_{t} \delta_{kt} Y_{tnn}
    \end{multlined} \label{eq:op_adj_weights_4}\\
    &= \sum_{\substack{k (n,m) \\ n<m}} w_{k (n,m)}
        \sum_{t} \delta_{kt} \left( Y_{tnn} + Y_{tmm} - Y_{tnm} - Y_{tmn} \right)
      \label{eq:op_adj_weights_5}\\
    &= \left\langle \bm W, \bm \Delta \underbrace{
        \big[ Y_{tnn} + Y_{tmm} - Y_{tnm} - Y_{tmn} \big]_{k (n,m)}
      }_{= d \bmr Y } \right\rangle \\
    &= \left\langle
      \bm W, \bmr L^* \left( \bmr Y, \bm \Delta \right)
    \right\rangle
    ,
  \end{align}
\end{subequations}
where between \cref{eq:op_adj_weights_4} and \cref{eq:op_adj_weights_5} we split the second sum into two sums over directed edges
\begin{equation*}
  \sum_{k(n,l)} = \sum_{k (n,l),\, n < l} + \sum_{k (n,l),\, l < n}.
\end{equation*}

On the other hand, for the adjoint operator with respect to coefficients, $\bmr L^{**}: \R_+^{K \times E} \times \R^{T \times N \times N} \to [0,1]^{K \times T}$, we can follow the previous derivation until \cref{eq:op_adj_weights_5}, then isolate the coefficient matrix and obtain
\begin{equation}\label{eq:op_adj_coeff}
  \left\langle \bm \Delta, \bmr L^{**} \left( \bm W, \bmr Y \right) \right\rangle
    = \left\langle \bm \Delta, \bm W d \bmr Y^\top \right\rangle
    .
\end{equation}

%% file: sections/content/algo_bi_pds.tex
\begin{algorithm}[t]
  \caption{\label{algo:bi-pds}
    Bilinear Primal-Dual Splitting (BiPDS)
  }
  \KwIn{Step parameters $\tau_w, \tau_c, \sigma \geq 0$, initial estimates $(\bm W_0, \bm \Delta_0, \bm Y_0)$, maximum number of iterations $N_{max}$}
  \KwOut{Estimated solution of \cref{eq:objective-nll-split}}

  \For{$n = 0$ \KwTo $N_{max}-1$}{
    $\bm W_{n+1} \gets
      \underset{\tau_w g_w}{\prox} \left( \bm W_n - \tau_w \left(
      \bmr L^{**}\left( \bm \Delta_n, \bm Y_n \right)
      + \nabla f_w (\bm W_n)
      \right) \right)
      $\nllabel{eq:bi-pds-w}\;
    $\bm \Delta_{n+1} \gets
      \underset{\tau_c g_c}{\prox} \left( \bm \Delta_n - \tau_c \left(
      \bmr L^*\left( \bm Y_n, \bm W_n \right)
      + \nabla f_c(\bm \Delta_n)
      \right) \right)
      $\nllabel{eq:bi-pds-delta}\;
    $\bm Y_{n+1} \gets \underset{\sigma h^*}{\prox} \left(
      \bm Y_n + \sigma
      \bmr L \left(
        2 \bm \Delta_{n+1} - \bm \Delta_n,
        2 \bm W_{n+1} - \bm W_n
      \right)
    \right)
    $\nllabel{eq:bi-pds-dual}\;

    \If{converged}{
      \Break
    }
  }
  \Return{$\bm W_{n+1}, \bm \Delta_{n+1}$}
\end{algorithm}

%% file: sections/experiment_model.tex
\section{Practical Models}\label{sec:models}

In this section we design some loss functions that we can optimize using \cref{algo:bi-pds}.
We leverage the most common hypotheses from graph structure learning and sparse representation literature to expand on the negative log-likelihood of the general model from \cref{eq:objective-nll}:
\begin{equation*}
  \underset{\bm W, \bm \Delta}{\arg\min}\
  \ell_w(\bm W) + \ell_c (\bm \Delta)
  + \ell_x(\bm X, \bm W, \bm \Delta))
  ,
\end{equation*}
then we explicitly provide the terms for \cref{eq:objective-nll-split} and derive their update steps for BiPDS.
We conclude with a complexity analysis of these models and their optimization with \cref{algo:bi-pds}.

\subsection{Objective function}\label{ssec:objective}

We start by defining the following base objective, with regularization hyperparameters $\alpha$'s:
\begin{subequations}\label{eq:objective-exp}
\begin{align}
  \ell_w (\bm W) &=
    \begin{multlined}[t]
      \chi_{\bm W \geq 0} + \alpha_{w L_1} \norm{\bm W}_1 + \\
      + \alpha_\perp \sum_{k' \neq k}^K \langle
        \bm w_k, \bm w_{k'}
      \rangle ,
    \end{multlined}
    \label{eq:ell-w} \\
  \ell_c (\bm \Delta) &=
    \chi_{\bm \Delta \in [0,1]} + \alpha_{c L_1} \norm{\bm \Delta}_1
    \label{eq:ell-c} , \\
  \ell_x (\bm \Delta, \bm W) &= \sum_{t=1}^T \left(
      \bm x_t^\top \bm L(\bm \delta_t, \bm W) \bm x_t
      + h (\bm L(\bm \delta_t, \bm W))
    \right)
    . \label{eq:ell-x}
\end{align}
\end{subequations}
The first terms in both $\ell_w$ and $\ell_c$ include the domain constraints on $\bm W$ and $\bm \Delta$, expressed as characteristic functions $\chi_{\R_+^{K \times E}}$ and $\chi_{[0,1]^{K \times T}}$ respectively, with
\begin{equation}
  \chi_A(\bm x) = \begin{cases}
    0 & \bm x \in A; \\
    +\infty & \bm x \notin A.
  \end{cases}
\end{equation}

The second terms of \cref{eq:ell-w,eq:ell-c} are weighted $L_1$-norms that promote sparsity in the atom weights and in their coefficients respectively.
They arise from $n$-variate exponential priors \citep{egilmezGraphLearningData2017} on the entries of $\bm W$ and $\bm \Delta$,
\begin{equation}\label{eq:exponential-prior}
  P(\bm y) = \left( \frac{\alpha}{2} \right)^n \exp\left(
    -\frac{\alpha}{2} \bm 1^\top \bm y
  \right)
  ,
\end{equation}
with $\bm y$ representing either atom weights or coefficients, and $n$ being the corresponding number of entries.
The third term in~$\ell_w$ penalizes dot products across atom weights, which promotes orthogonality and, since we also require the matrix to have nonnegative entries, this regularization encourages edges to be zero if they are already present in another atom.

In \cref{eq:ell-x}, the first term in the sum measures total variation of each signal on the Laplacians of instantaneous graphs, and is minimized by structures on which signals are smooth.
It corresponds to defining the prior signal distribution from \cref{eq:noise-filter} on the graph filter $k(\bm \Lambda) = \sqrt{\bm \Lambda^\dagger}$, with~$\dagger$ denoting the Moore-Penrose pseudo-inverse.
This signal distribution is known as Laplacian-constrained Gaussian Markov Random Field (LGMRF) \citep{egilmezGraphLearningLaplacian2016}:
\begin{equation}\label{eq:lgmrf-prior}
\left(\bm{x}_t \mid \bm{L}_t\right)
  = \frac{\exp \left(
    -\frac{1}{2} \bm{x}_t^{\top} \bm{L}_t \bm{x}_t
  \right)}{\sqrt{
    (2 \pi)^{N} \det_+ \left(\bm{L}_t^{\dagger}\right)
  }}
  ,
\end{equation}
where $\det_+$ represents the pseudo-determinant \citep{holbrookDifferentiatingPseudoDeterminant2018}.

Finally, the function $h$ describes additional hypotheses on instantaneous Laplacians, and in this work we propose two variants which give rise to the \emph{GraphDictLog} and \emph{GraphDictSpectral} models,
which share the same objectives for $ell_w$ and $ell_c$, and signal smoothness.
The \emph{GraphDictLog} model enforces node degrees to be positive for each instantaneous graph $G_t$ using a log barrier on the diagonal of the Laplacians, as done by \citet{kalofoliasHowLearnGraph2016} and derivative works:
\begin{equation}\label{eq:log-barrier-lapl}
  h(\bm L_t) =
    \bm 1_N^\top \log \left( \diag \left( \bm L_t \right) \right)
    .
\end{equation}
This formulation is more popular than using the logarithm of $\det_+(\bm L_t)$, which arises from \cref{eq:lgmrf-prior}, as it is more computationally efficient and less prone to numerical issues.

The \emph{GraphDictSpectral} model introduces a constraint that enforces all Laplacians to share the same eigenvectors $\bm U$, so that atoms represent filters on the same graph, instead of independent structures, since $k(\bm L_t) = \bm U k(\sum_k \delta_{kt} \bm \Lambda_k) \bm U^\top$.
This formulation corresponds to the model proposed by \citet{thanouParametricDictionaryLearning2013}.
We initialize $\bm U$ with the eigendecomposition of the empirical covariance matrix \citep{navarroJointInferenceMultiple2022}, so that the optimization is only performed on atoms eigenvalues and coefficients.
The specific $h$ term enforces this constraint with a characteristic function $\chi$, and includes the log barrier on the Laplacians pseudo-determinants \citep{holbrookDifferentiatingPseudoDeterminant2018} from the LGMRF prior:
\begin{equation}\label{eq:ell-spectral}
  h(\bm L_t) =
    \chi_{\bm {U \Lambda_t U}^\top} ( \bm L_t )
    + \log {\det}_+ ( \bm L_t )
    .
\end{equation}

\cref{eq:ell-w,eq:ell-c,eq:ell-x} are all sums of proximable and differentiable functions, and thus are compatible with \cref{eq:objective-nll-split}.


\input{sections/content/implementation.tex}


\subsection{Complexity analysis}\label{ssec:complexity}

We study the computational complexity of optimizing \emph{GraphDictLog} and \emph{GraphDictSpectral} with \cref{algo:bi-pds}, with respect to the number of nodes $N$, the number of atoms $K$, the number of samples $T$, and the maximum number of iterations of BiPDS $M$.

Starting from \emph{GraphDictLog}, we observe that the partial derivatives of $f$, in \cref{eq:grad-w,eq:grad-c} are dominated by matrix multiplications of complexity $\bigO (TKE)$, where $E$ is the number of edges, which is in $\bigO (N^2)$.
Computing the partial adjoint operators from \cref{eq:adjoint-op-degree} is in $\bigO (TNEK)$.
As the proximal operators in \cref{eq:prox-gw,eq:prox-ga} have complexity of $\bigO (KE)$ and $\bigO (TK)$ respectively, the updates of both $\bm W_n$ and $\bm \Delta_n$ are dominated by the adjoint computation from the dual variable.
In the dual update, we have the bilinear computation $\bm Y = \bm D \bm W^\top \bm \Delta^\top$ which has complexity of $\bigO (TKNE)$, and the proximal update from \cref{eq:prox-log} which is in $\bigO (NT)$.
Therefore, each iteration of \cref{algo:bi-pds} for optimizing \emph{GraphDictLog} parameters has a complexity of $\bigO (TNEK)$, which cumulates to $\bigO (MTNEK)$, or $\bigO(MTN^3K)$ for dense graphs.

For \emph{GraphDictSpectral}, we can leverage the fixed eigenvectors to reformulate the problem in the spectral space and greatly reduce computational complexity. More precisely, by taking the graph Fourier transform $\hat{\bm X} \bm U^\top$ of signals $\bm X$, we can reformulate \cref{eq:ell-x} to only depend on the sparse diagonal matrix of Laplacians eigenvalues.
This allows us to avoid computing eigendecompositions at each iteration, and therefore complexity is again dominated by computing the adjoint from the dual space to weights.
The overall computational cost of \cref{algo:bi-pds} for \emph{GraphDictSpectral} is again $\bigO (MTNEK)$.

%% file: sections/content/implementation.tex
\subsection{Implementation}\label{sec:implementation}

In this section we present the proximal operators and gradients corresponding to the GraphDict objective proposed above, namely in \cref{eq:ell-w,eq:ell-c,eq:ell-x,eq:log-barrier-lapl,eq:ell-spectral}.
In short, signals and dictionaries are bound by the smoothness term in \cref{eq:ell-x}, then we promote sparsity by L1 regularization for both coefficients and weights, and for the latter we additionally introduce an orthogonality constraint.
We regularize instantaneous Laplacians with either the log barrier on their diagonal from \cref{eq:log-barrier-lapl}, or with the log pseudo-determinant from \cref{eq:ell-spectral}

We start by providing an alternative formulation for the smoothness term in \cref{eq:ell-x}
\begin{align}
  \sum_{t=1}^T &
    \bm x_t^\top \bm L(\bm \delta_t, \bm W) \bm x_t
    = \sum_{t,k} \delta_{tk} \bm x_t^\top \bm L_{\bm w_k} \bm x_t
      \notag \\
    &= \sum_{t,k} \sum_{n=1}^N \sum_{m=1}^N
      \delta_{tk} x_{tn} x_{tm} \left[ \bm L_{w_k} \right]_{nm}
      \notag \\
    &= \sum_{t,k} \sum_{n=1}^N \delta_{tk} \left(
        \sum_{l \neq n}^N x_{tn}^2 w_{k (n,l)}
        - \sum_{m \neq n}^N x_{tn} x_{nm} w_{k (n,m)}
      \right) \notag \\
    &= \sum_{t,k} \sum_{(n,m) \in E}
      \delta_{tk} w_{k (n,m)} \left( x_{tn} - x_{tm} \right)^2
      \notag \\
    &= \norm{\bm{\Delta W} \odot \bm Z}_1
    ,
\end{align}
where in the last line we introduce the matrix $\bm Z \in \R^{T \times E}$ of squared pairwise differences over edges, with entries $[\bm Z]_{t (n,m)} = \left( x_{tn} - x_{tm} \right)^2$.
For the second-to-last equality, we split the sum over pairs of nodes into two sums over directed edges.
We remark that this formulation is equivalent to the one proposed by \citet{kalofoliasHowLearnGraph2016} to express signal smoothness over a single graph through pairwise node distances.

Putting together \cref{eq:ell-w,eq:ell-c,eq:ell-x} with the smoothness formulation above, our full objective for the experimental setup is
\begin{equation}
  \arg\min_{\bm W, \bm \Delta} g_w (\bm W) + g_c (\bm \Delta) + f(\bm W, \bm \Delta)
    + h \left( \bmr L(\bm \Delta, \bm W) \right),
\end{equation}
where
\begin{equation}
  \begin{aligned}
    g_w (\bm W) &= \chi_{\bm W \geq 0} + \alpha_{w L_1} \norm{\bm W}_1 \\
    g_c (\bm \Delta) &=
      \chi_{\bm \Delta \in [0,1]} + \alpha_{c L_1} \norm{\bm \Delta}_1 \\
    f(\bm W, \bm \Delta) &=
      \norm{\bm{\Delta W} \odot \bm Z}_1
      + \alpha_\perp \sum_{k' \neq k}^K \langle \bm w_k, \bm w_{k'} \rangle
      ,
  \end{aligned}
  \label{eq:experiment-loss}
\end{equation}
where $g_w$ and $g_c$ are proximable and $f$ is sub-differentiable.
The gradients of $f$ with respect to our variables ${\bm W \in \R_+^{K \times E}}$ and ${\bm \Delta \in \R^{T \times K}}$ are respectively
\begin{align}
  \nabla_{\bm W} f (\bm W, \bm \Delta) &=
    \bm{\Delta^\top Z} + \alpha_\perp \bm (\bm 1 \bm 1^\top - \Id_{K}) \bm W
    , \label{eq:grad-w} \\
  \nabla_{\bm \Delta} f (\bm W, \bm \Delta) &=
    \bm{Z W}^\top
    . \label{eq:grad-c}
\end{align}
The proximal operators of $g_w$ and $g_c$ are
\begin{align}
  \prox_{\tau_w g_w}(\bm W) &= \left( \bm W - \tau_w \alpha_{w L_1} \right)_+
    , \label{eq:prox-gw}\\
  \prox_{\tau_c g_c}(\bm \Delta) &= \left( \bm \Delta - \tau_c \alpha_{c L_1} \right)_{[0,1]}
    , \label{eq:prox-ga}
\end{align}
where in \cref{eq:prox-gw} the fuction ${(\bm x)_+ = \max(\bm x, 0)}$ is applied element-wise, as in \cref{eq:prox-ga} for ${(\bm x)_{[0,1]} = \max(\min(\bm x, 1), 0)}$.

For the function~$h$, we have two different settings, for which we compute the Fenchel conjugate~$h^*$.
In the \emph{GraphDictLog} case, we use the function from \cref{eq:log-barrier-lapl}, which sums the logarithms of node degrees for each instantaneous graph.
For this setting, we do not need to pass through the Laplacian, as we only need the degree vectors of instantaneous graphs, which we can compute with the linear operator $\bm D \in \R^{N \times E}$ defined as
\begin{equation}
  \left[ \bm{D w} \right]_n = \sum_{m \neq n} w_{(n,m)}
\end{equation}
The dual term is therefore $\bm Y = \bmr D(\bm \Delta, \bm W) = \bm D (\bm{\Delta W})^\top \in \R^{N \times T}$, for which we have the two partial adjoints
\begin{equation}
\begin{aligned}
  \bmr D^* (\bm Y, \bm W) &= \bm Y^\top \bm D \bm W^\top , \\
  \bmr D^{**} (\bm \Delta, \bm Y) &= \bm\Delta^\top \bm{Y D}.
\end{aligned}
\label{eq:adjoint-op-degree}
\end{equation}
Finally, the proximal operator of the Fenchel conjugate of $h$ is
\begin{equation}
  \prox_{\sigma h^*} (\bm Y) = \frac{\bm Y - \sqrt{\bm Y^2 + 4 \sigma}}{2}
  . \label{eq:prox-log}
\end{equation}

On the other hand, for \emph{GraphDictSpectral}, we use the main formulation for $\bmr Y = \bmr L(\bm \Delta, \bm W)$ and the function $h$ from \cref{eq:ell-spectral}. Fixing the eigenvectors for all Laplacians to be $\bm U \in \R^{N \times N}$, the proximal operator of $h^*$ is
\begin{equation}
  \left[ \prox_{\sigma h^*}(\bmr Y) \right]_t =
    \bm U \left(
      \frac{ \bm \Lambda_t - \sqrt{\bm \Lambda_t^2 + 4 \gamma \Id_N} }{2}
    \right) \bm U^\top
    , \label{eq:prox-logdet-star}
\end{equation}
where, $\bm \Lambda_t$ are the eigenvalues each \emph{instantaneous Laplacian}~$\bm Y_t$ in the tensor~$\bmr Y$; the square root is applied component-wise on the diagonal and the identity is broadcasted as needed.

%% file: sections/experiments.tex
\section{Experiments}\label{sec:experiments}
We present three experiments to illustrate the capabilities of our GraphDict models.
First, we show their ability to reconstruct superposed graphs from synthetic signals, outperforming state-of-the-art clustering methods.
Second, we demonstrate their effectiveness in recovering time-varying graphs from sequential data, where they surpass dedicated temporal models.
Finally, we apply them to a real-world EEG dataset to learn explainable brain states for a motor imagery classification task, achieving superior performance with fewer features than methods from the neuroscientific literature.


\subsection{Superposing Graphs}\label{ssec:exp-clustering}


\subsubsection{Description}

In this experiment we generate data according to the Graph-Dictionary Signal Model from \cref{sec:graph-signal-model}, and we focus on recovering \emph{instantaneous} graphs.
In particular, we investigate the evolution of model performances with respect to the maximum number of atoms that can contribute to each signal, which we call \emph{superposition}.
With a superposition of one, each signal arises from a single network and the problem is equivalent to signal clustering and graph reconstruction.
Therefore, we compare GraphDictLog and GraphDictSpectral to \emph{Gaussian Mixtures}---where we estimate the graph as the pseudo-inverse of the empirical correlation---\emph{KGraphs} \citep{araghiGraphsAlgorithmGraph2019}, and \emph{GLMM} \citep{mareticGraphLaplacianMixture2020}, as these methods are state of the art for the latter problem, but no model exists for higher superpositions.
We re-implement the two latter, and we use Gaussian Mixtures from Scikit-learn \citep{pedregosaScikitlearnMachineLearning2011}.


\subsubsection{Data}\label{ssec:exp-clustering-data}

\begin{figure}[t]
  \centering
  \begin{tikzpicture}[every node/.style={align=center, font=\small}]
    \node (img) [inner sep=0pt] {\includegraphics[width=.9\linewidth]{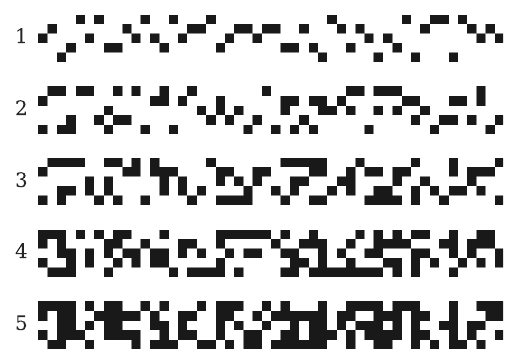}};
    \node (ylabel) at (img.west) [rotate=90, anchor=base] {Superposition};
    \path[->,semithick] (img.south west) + (10pt,0) edge node[below] {Sample Id} (img.south east);
  \end{tikzpicture}
  \caption{\label{fig:reconstruct-coeff}
    Samples of coefficient matrices for different \emph{superposition} values
    Each block represents in black the positive coefficients of five atoms over 50 samples.
  }
\end{figure}

We sample a dictionary of $K=5$ graphs from an Erd\"os-Renyi distribution with 30 nodes and $p=0.2$ edge probability (ER$(30, 0.2)$), which is shared by the following generation steps.
Then, we define the \emph{superposition} parameter $s \in \{1, \dots, 5\}$, that indicates the maximum number of atom coefficients that can be nonzero at the same time for each experiment run.
Subsequently, for each $t \in \mathbb{N}$, we build a random coefficient vector $\bm \delta_t$ by first sampling the number of positive coefficients uniformly from 1 to $s$, and then by randomly selecting the atoms.
\cref{fig:reconstruct-coeff} shows multiple sampled coefficients for each superposition value.
We observe that, by definition, at most $s$ atoms have nonzero coefficients, but some $\bm \delta_t$ have less positive values.

We gather multiple independent samples from this process as columns of two matrices of discrete coefficients $\bm \Delta_{tr}\in \{0,1\}^{5 \times T_{tr}}, \bm \Delta_{te} \in \{0,1\}^{5 \times T_{te}}$, and from the corresponding instantaneous graphs $\bm \Delta_{tr \mid te} \bm W$ we sample two sets of signals $\bm X_{tr}$ and $\bm X_{te}$ following a Laplacian-constrained Gaussian Markov Random Field, or LGMRF \citep{egilmezGraphLearningLaplacian2016}.
In this model the combinatorial Laplacian corresponds the precision matrix of a Gaussian distribution, so that signals follow \cref{eq:noise-filter} for the filter $k(\bm \Lambda) = \sqrt{\bm \Lambda^\dagger}$.
We use $\bm X_{tr}$ and $\bm X_{te}$ respectively for training and testing the models.
Each atom contributes to multiple instantaneous graphs as we allow for more superposition, so we balance the tasks by setting $T_{tr}$ to allow for each atom to contribute to 500 samples on average.
More precisely, for $s=1, \dots, 5$ we have $T_{tr}=2500, 1666, 1250, 1000, 833$; while the test set has a fixed size of $T_{te} = 500$.

\begin{figure*}[t]
  \centering
  \subfloat{
    \includegraphics[height=0.3\linewidth]{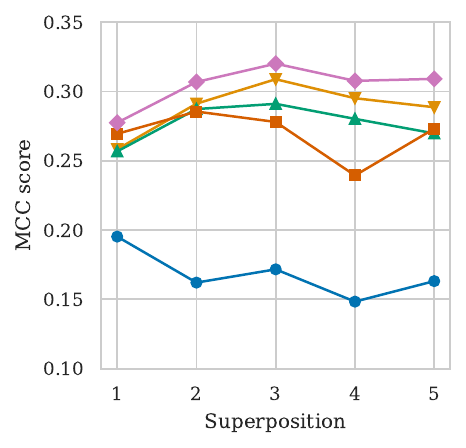}
  }
  \hfill
  \subfloat{
    \includegraphics[height=0.3\linewidth]{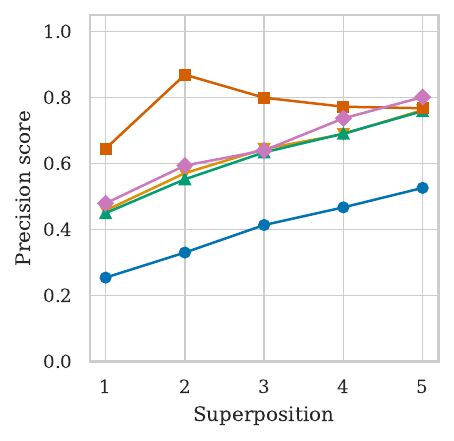}
  }
  \hfill
  \subfloat{
    \includegraphics[height=0.3\linewidth]{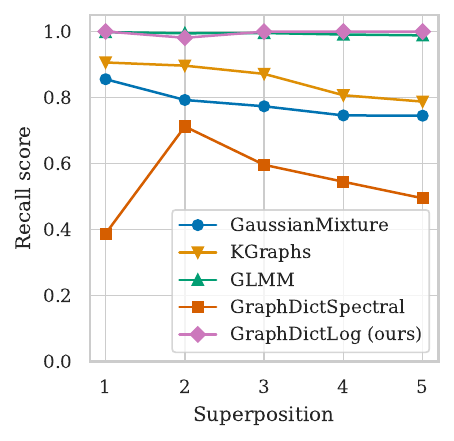}
  }
  \caption{\label{fig:reconstruct-mcc-trp}
    Test performance on edge recovery of instantaneous graphs from signals, measured by Matthews correlation coefficient (MCC), precision and recall.
    Results are averaged over five random seeds.
  }
\end{figure*}


\subsubsection{Results}

We focus on the ability of models to correctly recover edges, and we measure their performances with the Matthews correlation coefficient (MCC) score:
\begin{equation}\label{eq:mcc}
  MCC = \frac{tp \times tn - fp \times fn}{\sqrt{(tp + fp)(tp + fn)(tn + fp)(tn + fn)}}
  ,
\end{equation}
which takes into account true ($t$) and false ($p$) positive ($p$) and negative ($n$) edges.
This metric, which takes values in the $[-1,1]$ interval, is better suited than the F1 score in comparing edges presence or absence in sparse graphs, thanks to being balanced even with classes of very different sizes.
For every graph-learning method, we perform a hyperparameter grid search by training on $\bm X_{tr}$ and scoring on the instantaneous graphs of the same training set.

\cref{fig:reconstruct-mcc-trp} shows the evolution of the MCC, precision, and recall scores on the test set for increasingly superposed coefficients, averaged on five different random seeds.
For a superposition value of 1, which corresponds to signal clustering, all graph learning methods perform similarly, while Gaussian Mixtures lag behind.
With increasingly superposed graphs, GraphDictLog consistently provides the best results in terms of MCC.
The overall high recall shows that all models, except GraphDictSpectral, recover too many edges. Furthermore, all precision curves increase with more atoms being superposed, in spite of the higher degrees of freedom of instantaneous graphs.
Focusing on our GraphDictLog model, we observe that in this setting, with the low edge probability in the ER graphs, the most useful prior is sparsity on atom weights, from \cref{eq:ell-w}, controlled by~$\alpha_{w L_1}$.


\begin{table*}[t]
  \caption{\label{tab:time-series}
    Edge recovery performances, measured by average MCC score across all instantaneous graphs, for the time series experiments. Each stochastic model is randomly initialized multiple times, and we report its average score and standard deviation.
  }
  \begin{center}
  \scalebox{.915}{
  \input{tables/ts_results_mean_only.tex}
  }
  \end{center}
\end{table*}

\subsection{Time-Varying Graphs}\label{ssec:exp-time}


\subsubsection{Description}

In this experiment, we observe signals sequentially, as time series, and we aim to recover the underlying instantaneous graphs, which have temporal dependencies.
Multiple models have been proposed for these settings, introducing priors on the evolution of networks over time, and we can have a fair comparison with three different benchmarks.
First, \emph{WindowLogModel} is a temporal version of the static graph-learning method from \citet{kalofoliasLargeScaleGraph2018}, which learns independent graphs over sliding windows.
\emph{TGFA} is a dynamic graph-learning paradigm encompassing the models from \citet{kalofoliasLearningTimeVarying2017} and \citet{yamadaTimeVaryingGraphLearning2020}, which adds an L1 or L2 regularization on edge changes between adjacent windows to the WindowLogModel objective.

\citet{yamadaTemporalMultiresolutionGraph2021} take a different approach and suppose that the time-varying graphs can be represented in a hierarchical manner.
They organize windows as a binary tree, with neighbors sharing parents, and assign learnable graphs to this hierarchy.
The signal distribution within each window is characterized by an instantaneous network given by the sum of all graphs from the root to the corresponding leaf, so that the hierarchical graphs can be estimated with a maximum likelihood formulation.
Due to its additive nature, this model falls within our dictionary framework, with graphs in the hierarchy corresponding to atoms, but with a substantial difference in the coefficients.
In their case, these are fixed and describe the tree structure, while in the dictionary learning setting they are parameters of the model.
We implement this hierarchical graph-learning model as \emph{GraphDictHier}, by fixing the atoms coefficients and only learning atom edge weights.

For this experiment, we include a temporal prior in GraphDictLog and GraphDictSpectral, similarly to TGFA, by introducing an $L_1$ regularization on the difference in coefficients of adjacent windows to \cref{eq:ell-c} to promote sparsity in changes
\begin{equation}\label{eq:ell-c-time}
  \alpha_{\mathrm{diff}} \sum_{t=1}^{T-1} \norm{\bm \delta_{t+1} - \bm \delta_t}_1
  .
\end{equation}
This term encourages atom coefficients to be constant between $t$ and $t+1$, or to change abruptly.
We also allow for coefficients to be fixed over windows, with the size becoming a hyperparameter with the same search space as benchmarks.


\subsubsection{Data}

We generate sequential graphs from two time-varying distributions, which we use in separate runs:
\begin{itemize}
  \item The \emph{Edge-Markovian evolving graph} (EMEG) model starts with a sample of ER$(36, 0.1)$, and at each graph change we add new edges or remove existing ones from the current network with probabilities of 0.001 and 0.01 respectively.
  \item The \emph{Switching behavior graph} (SBG) model consists of six independent ER$(36, 0.05)$ networks, that define states in a Markov Process. At each step we either keep the same graph as the previous one, with probability of 0.98, or change uniformly to another one.
\end{itemize}
In both models, weights for new edges are sampled uniformly on the $[0.1,3]$ interval, and do not change until the edge disappears.
With these models we obtain sequences of $T_G$ weights $W \in \R_+^{T_G \times E}$,
that allow us to sample signals from the LGMRF distribution presented in \cref{ssec:exp-clustering-data}.
For each time-varying graph process, we study two signal generation settings:
\begin{enumerate}
  \item Graphs are stable over windows: we set $T_G = 32$, and we sample $W=20$ consecutive signals $\bm X_w \in \R^{W \times N}$, for a total of 640 measurements;
  \item The system is continually evolving: we sample $T_G = 512$ graphs each of which produces a single signal $\bm x_t \in \R^{N}$.
\end{enumerate}


\subsubsection{Results}

\cref{tab:time-series} shows the performances in edge recovery of benchmarks and the different GraphDict models on all time series tasks, averaged over five random seeds. We select hyperparameters of every model by grid search and evaluation on training reconstruction.
Including the additional regularization from \cref{eq:ell-c-time} in the loss term proved to be an effective prior for the EMEG process; while enforcing coefficients to change over windows instead of over samples improved reconstruction performances for the SBG process.

We see that our GraphDictLog greatly outperforms other methods in all settings, showing that it can capture the gradual evolution of the EMEG distribution, as well as the recurrent graphs of SBG. In particular, it achieves the highest precision in all scenarios, indicating that the recovered edges are more likely to be correct. However, the model exhibits a lower recall compared to other methods, which might suggest that the sparsity priors on weights and coefficients were too strong, despite hyperparameter tuning.
We see that the scores of all models are consistently higher in the stable $32 \times 20$ setting, in particular thanks to the hypothesis of window-based models being correct.
We suppose that the advantage over the hierarchical model from \citet{yamadaTemporalMultiresolutionGraph2021}, which they presented as state of the art, comes from the greater flexibility of our model in learning on its own when atoms contribute to a sample, or window. In particular, for the SBG setting, in which states are recurrent, our model can ``reuse'' atoms across time, while the hierarchical model has less memory across windows far apart.
This also explains why GraphDictHier shows a higher recall than GraphDictLog, as the consistency of neighboring samples might enforce to focus on fewer persistent edges.


\subsection{Brain States For Motor Imagery}\label{ssec:exp-brain}


\subsubsection{Description}

In this final experiment we study the capability of GraphDict to learn an explainable representation for a downstream classification task, in particular in its GraphDictLog form.
We analyze brain activity measurements  with the objective of disambiguating left and right-hand motor imagery.
Neurologists often resort to \emph{microstates} \citep{michelEEGMicrostatesTool2018} to characterize brain signals, and they obtain such states by clustering EEG signals over time.
This is an interesting setting for our model, where instead of single microstates we can reconstruct a dictionary of states (graph atoms) that can provide insights on brain connectivity.

We infer brain states as the clusters, or atoms, learned by GraphDictLog and three baseline methods:
\emph{KMeans}, which is the standard method for microstate inference; \emph{Gaussian Mixtures}, which allow for soft assignments \citep{mishraEEGMicrostatesContinuous2020}; \emph{GLMM}, that include network properties on states \citep{ricchiDynamicsFunctionalNetwork2022}.
We perform the classification task on separate time windows, called \emph{events}.
For each event we compute three features for each state, based on the evolution of attribution probabilities, for clustering methods, and coefficients, for graph dictionaries:
\begin{itemize}
  \item \emph{Number of occurrences} (OCC): count of continuous intervals covered;
  \item \emph{Coverage} (COV): total time over the window;
  \item \emph{Average duration:} average length of continuous occurrences in seconds, i.e. OCC/COV.
\end{itemize}


\subsubsection{Data}

The data for this experiment consists in a set of annotated Electroencephalography (EEG) signals from the EEG BCI dataset \citep{schalkBCI2000GeneralpurposeBraincomputer2004}, provided by the MNE-Python library \cite{gramfortMEGEEGData2013}.
EEGs are multivariate time series of electrical potentials captured by electrodes placed over the subjects scalp.
This dataset collects measurements from 15 subjects, each with 45 \emph{events} spanning the onset of imagined motion of either hand, with 23 for the left and 22 for the right hand.


\begin{figure}[t]
    \includegraphics[width=\linewidth]{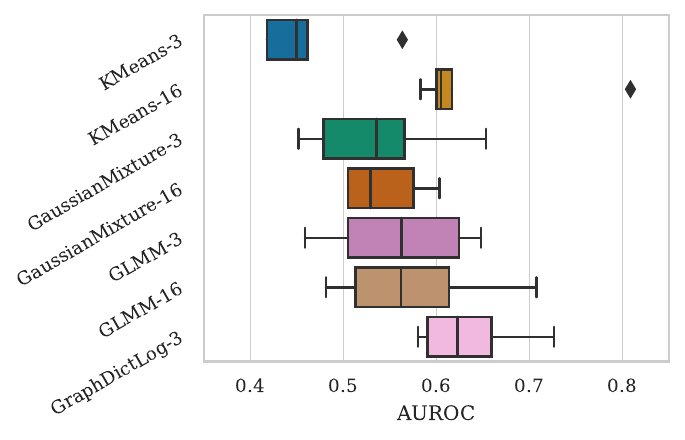}
    \caption{\label{fig:brain-forest-cv}
      Distribution of test scores for motor imagery classification from brain state features, computed by leave-one-subject-out cross-validation.
      On the y-axis we find the name of the model used for learning brain states, together with the number of clusters, or atoms, defining such states.
    }
\end{figure}


\subsubsection{Pipeline}

\input{sections/content/exp_details.tex}


\subsubsection{Results}

\begin{figure}[t]
  \centering
  \input{figures/brain/brain_atoms.tex}
  \caption{\label{fig:brain-atoms}
    Atoms and 50 corresponding coefficients learned by \texttt{GraphDictLog} on EEG signals from motor imagery data. Atoms are sorted by frequency of appearance and show edges between electrodes. The coefficient matrix is color coded from zero, in white, to one, in black, and has atoms as rows and sample indices as columns.
  }
\end{figure}

\cref{fig:brain-forest-cv} shows the results' distribution of random forest classifiers on 15-fold cross validation, with each fold corresponding to a unique subject. For each method we test different numbers of states, or clusters, and select other hyperparameters to obtain the highest average training performance in classification.
We see that GraphDictLog with three atoms obtains consistently the highest AUROC, with the only comparable model being KMeans with 16 clusters. Still, KMeans-16 requires 48 features, against the nine features arising from the three states of GraphDictLog-3.
Interestingly, of all regularization terms from \cref{eq:ell-w,eq:ell-c,eq:ell-x}, only $\alpha_{\perp}$, which enforces edge orthogonality, is positive.

In this setting, we observe a strong influence of the physics of the problem on the data, which translates in electrodes being close in space to be highly correlated due to the skull acting as a diffuser for brain activity.
To avoid this strong correlation to take over, enforcing atom orthogonality proves to be an effective solution.
The top of \cref{fig:brain-atoms} shows the reconstructed atoms corresponding to the best classification model, ordered by their average activation from low to high.
We see many edges between electrodes close in space, but the atoms seem to capture functional areas, as we recognize the frontal lobe in Atom 1, and the increased connectivity of the occipital lobe in Atom 2 might correspond to visual activity.
The bottom part of the figure shows the coefficients of the signals from 50 peaks of GFP, of which the first 28 are from the imagined movement of the right hand, and the rest from the left hand.
Even though we cannot identify any meaningful pattern in the coefficients, we can examine feature importance in the random forest model, in terms of mean decrease in impurity.
Interestingly, models focus almost entirely on the coefficient features corresponding to Atoms~1 and~2, each of which gets almost equal importance ($\approx 0.16$).

%% file: tables/ts_results_mean_only.tex
\sisetup{
  detect-weight,
  mode=text,
  reset-math-version=false,
  uncertainty-mode=separate,
  table-format=2.1(1.1)
}
\setlength{\tabcolsep}{3pt}
\begin{tabular}{@{}l|SS[table-format=2.1(2.1)]S|SSS|SSS|SSS[table-format=2.1(2.1)]|}
\toprule
Distribution & \multicolumn{6}{c|}{EMEG} & \multicolumn{6}{c|}{SBG} \\
\cmidrule(rl){2-7} \cmidrule(rl){8-13}
graphs $\cdot$ win.~size & \multicolumn{3}{c|}{$32 \cdot 20$}& \multicolumn{3}{c|}{$512 \cdot 1$} & \multicolumn{3}{c|}{$32 \cdot 20$} & \multicolumn{3}{c|}{$512 \cdot 1$} \\
\cmidrule(rl){2-4} \cmidrule(rl){5-7} \cmidrule(rl){8-10} \cmidrule(rl){11-13}
Metric & {MCC} & {Prec} & {Rec} & {MCC} & {Prec} & {Rec} & {MCC} & {Prec} & {Rec} & {MCC} & {Prec} & {Rec} \\
\midrule
\ttfamily WindowLogModel & 37.7 & 35.6 & 55.7 & 36.0 & 37.9 & 51.2 & 33.6 & 22.6 & 66.9 & 28.6 & 16.4 & 72.3 \\
\ttfamily TGFA & 20.7 & 16.5 & 63.0 & 17.8 & 15.4 & \best 86.5 & 14.0 & 8.7 & 69.7 & 15.7 & 8.5 & 74.1 \\
\ttfamily GraphDictHier & 50.6(0.4) & 36.6(1.2) & \best 87.8(2.1) & 40.2(0.9) & 30.8(1.4) & 80.8(1.9) & 32.8(2.1) & 19.7(1.9) & 76.5(0.8) & 31.2(0.3) & 17.9(0.1) & \best 75.5(0.2) \\
\ttfamily GraphDictSpectral & 57.2(6.7) & 52.8(11.6) & 71.6(3.3) & 26.9(0.7) & 40.2(3.3) & 27.7(4.6) & 29.6(1.5) & 16.3(0.5) & \best 81.6(3.3) & 18.4(0.7) & 22.3(6.6) & 27.1(12.0) \\
\ttfamily GraphDictLog & \best 61.0(1.6) & \best 86.2(1.8) & 46.7(2.5) & \best 46.2(2.0) & \best 58.4(4.0) & 43.9(4.0) & \best 48.1(1.1) & \best 40.7(1.4) & 64.7(0.6) & \best 44.0(2.2) & \best 39.8(3.0) & 52.1(2.7) \\
\bottomrule
\end{tabular}

%% file: sections/content/exp_details.tex
We preprocess the data according to microstates' literature \citep{michelEEGMicrostatesTool2018, mishraEEGMicrostatesContinuous2020} to obtain clean signals for the 64 EEG channels, from which we infer brain states and their relative features.
The full pipeline consists in the following five steps, which we perform separately for every benchmark clustering model and our graph-dictionary learning one, with multiple hyperparameters each:
\begin{enumerate}
  \item \textbf{Data preprocessing} according to Microstates literature:
  \begin{enumerate}
    \item We re-reference EEG signals and center each sample to the average over all electrodes
    \begin{equation}
      \bm x_t \gets \bm x_t - \frac{1}{N} \sum_{n=1}^N x_{tn} ;
    \end{equation}
    \item We apply a band-pass filter between \num{0.5} and \qty{30}{Hz};
 \end{enumerate}
  \item \textbf{State embedding}: we identify the \enquote{cleanest samples of brain activity}, which according to neurologists are peaks of the \emph{Global Field Power} (GFP), which we use to identify brain states. More precisely:
  \begin{enumerate}
    \item We compute Global Field Power, which is the standard deviation across channels of each sample;
    \item We identify the \emph{peaks} of this series;
    \item We train the given clustering or GraphDict model and retrieve cluster, or atom, assignments of peaks;
    \item We assign to each samples in EEG events the same coefficients as its closest peak in time;
  \end{enumerate}
  \item \textbf{Feature extraction}: we compute the pre-cited state features for each EEG event;
  \item \textbf{Classification}: we train and test multiple random forest classifiers with leave-one-patient-out cross validation;
\end{enumerate}

Finally, we gather the results for each state embedding model.
We report the distribution of test scores on motor imagery classification corresponding to the best hyperparameters with the best downstream classifier.

%% file: figures/brain/brain_atoms.tex
\begin{tikzpicture}[
  scale=1,
  every node/.style={align=center, font=\small, inner sep=0pt}
]
  \node (a0) at (-2.8cm,0cm) {\includegraphics[width=2.4cm]{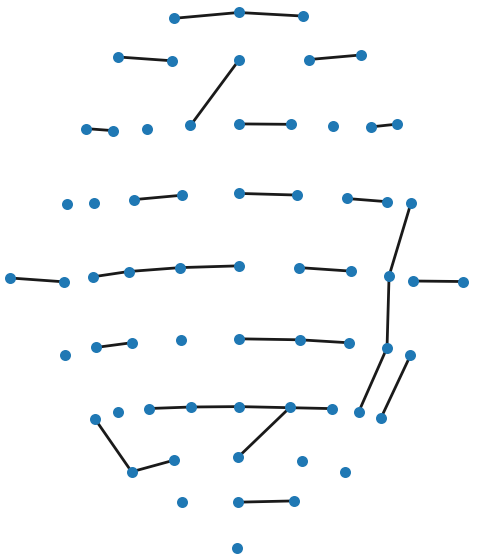}};
  \node (a1) at ( 0cm,0cm) {\includegraphics[width=2.4cm]{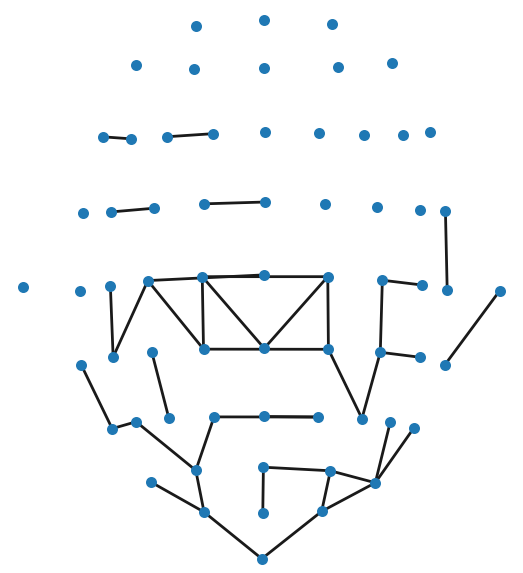}};
  \node (a2) at ( 2.8cm,0cm) {\includegraphics[width=2.4cm]{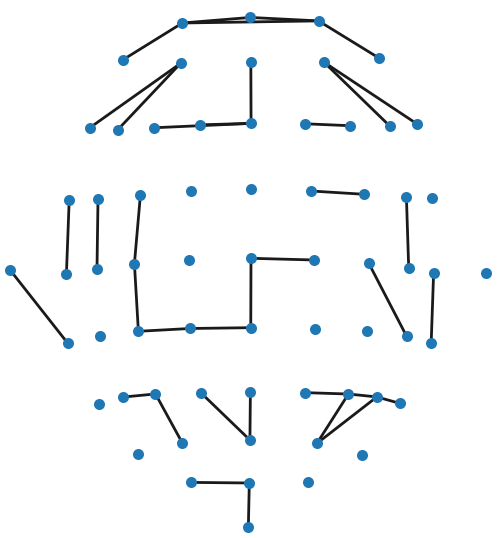}};

  \node[below=1ex of a0] {Atom 0};
  \node[below=1ex of a1] (label1) {Atom 1};
  \node[below=1ex of a2] {Atom 2};

  \node[below=1em of label1] (coeffs) {
  \includegraphics[width=8cm]{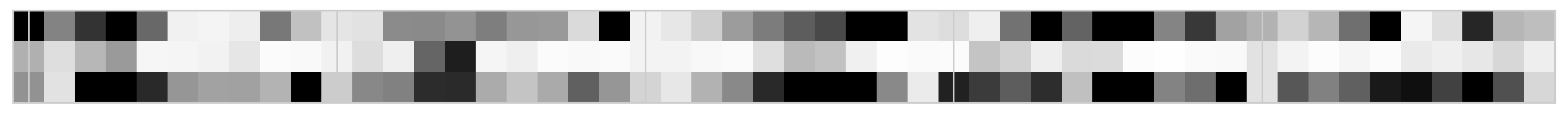}};
  \draw[->, thick] (coeffs.south west) -- (coeffs.south east);
  \node[below right=0.5ex of coeffs] {$T$};
  \node[left=0.5ex of coeffs] {$k$};
  \node[below=1ex of coeffs] {Coefficients};
\end{tikzpicture}

%% file: sections/conclusion.tex
\section{Conclusion}\label{sec:conclusion}

We propose \emph{GraphDict}, a Graph-Dictionary signal model for multivariate data, together with an optimization algorithm to jointly learn atoms and their coefficients.
This framework characterizes systems with complex dynamics and provides novel insights on data, by reconstructing a dictionary of graphs that defines the probability distribution of the observed phenomena. Furthermore, our model provides sparse representations in the form of atom coefficients, which are inherently explainable as they describe the influence of the corresponding graph on the data generating process.
Atom coefficients of our GraphDict model are also shown to be relevant in downstream tasks, for instance to describe brain states for motor imagery classification problems.

Our formulation can also leverage expert knowledge, which we include as priors on signals, weights, and coefficients.
In all applications, we observe that choosing well-thought priors significantly improves the model capabilities.
More precisely, weight sparsity is very helpful when it reflects the ground truth graphs, as in our synthetic settings.
For temporal data regularizing changes between activations over time helps capture the sequential aspect of the underlying process.

The expressive power of our framework comes with some limitations that open up several promising directions for future research.
A first set of challenges relates to the specific instantiation of our model presented in \cref{sec:models}. While the general formulation in \cref{eq:objective-nll} is highly flexible, our practical implementation relies on specific priors and regularization terms (e.g., $\alpha_{w L_1}, \alpha_{c L_1}, \alpha_\perp$) that require heuristic tuning. The high computational complexity, also tied to this specific model, makes extensive hyperparameter search challenging. This is not an inherent limitation of the GraphDict framework itself, but rather of the chosen objective function. A key direction for future work is to explore other models within our framework, potentially with different regularizers or data fidelity terms that are more computationally efficient or allow for a more principled theoretical analysis of their impact on the estimated parameters.

Second, while our proposed BiPDS algorithm performs well empirically, a formal proof of its convergence remains an open theoretical question. Establishing convergence guarantees, perhaps under specific assumptions on the objective function, would be a significant contribution that would bolster the reliability of our method and could inspire other solutions for similar bilinear problems.

The generality of our framework opens up numerous other exciting extensions. The current model assumes a static dictionary, but it could be extended to allow atoms to evolve over time. Similarly, while we used an LGMRF model, the framework can readily incorporate other graph signal models. An ambitious but promising direction is to integrate principles from optimal transport, as discussed in \cref{sec:related}, to handle signals from graphs with non-aligned or varying node sets. These future explorations underscore the broad applicability and potential of our core contribution.

Thanks to the flexibility of GraphDict and its optimization algorithm to accommodate for different regularization terms and priors, this framework can leverage domain knowledge and provide insightful representations of multivariate data from different domains.

%% file: sections/acknowledgments.tex
This work was supported by the SNSF Sinergia project \enquote{PEDESITE: Personalized Detection of Epileptic Seizure in the Internet of Things (IoT) Era} under grant agreement CRSII5 193813.